\documentclass[conference]{IEEEtran}
\IEEEoverridecommandlockouts

\usepackage{cite}
\usepackage{amsmath,amssymb,amsfonts}
\usepackage{algorithmic}
\usepackage{graphicx}
\usepackage{textcomp}
\usepackage{xcolor}
\usepackage{authblk}
\usepackage{float}
\usepackage{hyperref}
\def\BibTeX{{\rm B\kern-.05em{\sc i\kern-.025em b}\kern-.08em
    T\kern-.1667em\lower.7ex\hbox{E}\kern-.125emX}}
\begin{document}

\title{A Text-based Approach For Link Prediction on Wikipedia Articles}

\author[1,2,3]{Anh Hoang Tran}
\author[1,2,3]{Tam Minh Nguyen}
\author[1,2,3]{Son T. Luu}
\affil[1]{Faculty of Information Science and Engineering}
\affil[2]{University of Information Technology}
\affil[3]{Vietnam National University, Ho Chi Minh City, Vietnam}

\affil[ ]{Email: \textit {\{20521079, 20520748\}@gm.uit.edu.vn}, \textit{sonlt@uit.edu.vn}}
\affil[ ]{Kaggle team: \textit{DS$@$UIT\_SAT}}

\maketitle

\begin{abstract}
This paper present our work in the DSAA 2023 Challenge about Link Prediction for Wikipedia Articles. We use traditional machine learning models with POS tags (part-of-speech tags) features extracted from text to train the classification model for predicting whether two nodes has the link. Then, we use these tags to test on various machine learning models. We obtained the results by F1 score at 0.99999 and got $7^{th}$ place in the competition. Our source code is publicly available at this link: \url{https://github.com/Tam1032/DSAA2023-Challenge-Link-prediction-DS-UIT_SAT}. 
\end{abstract}


\section{Introduction}
Link prediction is a task that aims to predict the existence of connection between two nodes in the networks \cite{10.1145/3012704}. When the data is represented in the graph, link prediction can extract potential relationship between objects, which is useful for retrieving valuable information from large amount of data. According to \cite{LIU2023118737}, the link prediction is defined as given a finite set of nodes V and edges E as input, we must find a function T that map each pair of nodes (i,j) to a real number [0,1] as output, where i and j are not belong to E. In general, we can treat this task as a classifier task, specifically the binary classification. 

Wikipedia\footnote{\url{https://www.wikipedia.org/}} is the largest encyclopedia where the articles are bound together by the hyperlinks \cite{zesch-gurevych-2007-analysis}. By predicting future links between articles, we can enhance the navigability and discoverability of the network, and provide users with more relevant and informative articles through the links. The DSAA 2023 Competition focus on the link prediction task applied to Wikipedia articles. In this challenge, we are given a sparsified subgraph of the Wikipedia network, and our target is to predict if a link exists between two Wikipedia pages u and v. In particular, we are given a ground-truth file which contains pairs of nodes corresponding to positive or negative samples. If an edge exists between two nodes then the corresponding label is set to 1, otherwise, the label is 0. However, if a pair of nodes is not reported in the file, this does not imply that there is no edge between them. Some of these missing pairs of nodes will appear in the test file, and we will have to predict whether there is a link between them or not.

In this paper, we will present our approach and solutions for this challenge. Our approach is text-based, and we used the Part-of-Speech tagging (POS) to extract features from the text. Before running prediction models, we first analyzed and visualized the data to understand more about the dataset. Next, we embedded the nodes by applying POS tagging, and we also conducted statiscal t-test to select the tags. Finally, we ran the classification models on the embedded dataset. Most of the models we used are classical Machine Learning models, which ensures the efficiency of our approach. Our method archieved 0.99999 in both public and private test sets, and placed $7^{th}$ in the competion.

\section{Related Works}
The methods used for link prediction in early works include similarity-based methods \cite{ADAMIC2003211}, maximum likelihood models \cite{clauset2008hierarchical} and probabilistic models \cite{Getoor2001}. More recently, network embedding methods such as DeepWalk (DW) \cite{10.1145/2623330.2623732} and node2vec \cite{10.1145/2939672.2939754} have shown better performance in several applications. However, in document networks, these models cannot use the text content associated with the nodes. Therefore, we propose a text-based approach in this competition to utilize the textual features underlying each node.

According to \cite{DAUD2020102716}, supervised learning is one potential approach for the link prediction system. Since this task is categorized as binary classification, we can adapt several machine learning methods to the task such as Logistic Regression \cite{nigam1999using}, KNN \cite{fix1952discriminatory}, Tree-based algorithms like Decision Tree and Random Forest \cite{598994}. Besides the machine learning models, deep learing models such as ANN network also bring optimistic results compared to traditional machine learning models \cite{8365780}. Also, as shown in \cite{DAUD2020102716}, to improve the performance of model in large dynamic network, the hybrid solution is a suitable approach because it can combine multiple similarity features with an operation by the evolutionary algorithm. We chose the two famous models for this approach including LightBGM\cite{ke2017lightgbm} and XGBoost\cite{chen2016xgboost}. 

Besides with the machine learning models, the way to extract the features from text is also important since it capture the semantic meaning of text in the article for representing the potential link. The authors in \cite{van1991nlp} show that when combine the POS tag features with word features, the accuracy of text classification model is increased. Also, from the work in \cite{stoica2020improving} and \cite{badaro2020link}, the authors used POS tags as features to extract relevant information for the link prediction task. Consequently, we use the POS tags as feature in this paper to extract semantic information from articles for the link prediction task.
\section{Proposed Methodology}
For this challenge, we used Python programming language and code in Jupyter notebook (.ipynb) files in Google Colab. We used Machine Learning models which doesn't require much computing power, so we don't have to worry about running out of resources when training in Google Colab. Detail information about the libraries we used is provided in the table \ref{library} below:
\begin{table}[h]
\begin{center}
\caption{Python libraries.}
\label{library}%
\begin{tabular}{ll}
\textbf{Purpose} & \textbf{Python libraries}\\
\hline
Basic & Pandas, numpy, tqdm \\
Illustration & matplotlib, seaborn \\
Node embedding & nltk, re \\
Machine Learning libraries & sklearn, xgboost, lightgbm, tensorflow, keras \\
\hline
\end{tabular}
\end{center}
\end{table}
\subsection{Data exploration}
The Link Prediction for Wikipedia Articles dataset includes 2 main components: train.csv and node.tsv. The train.csv contains pair of nodes that has a link between them or not. The training file consists of 948,232 samples, including 512,389 with label 0 and 435,843 with label 1. It can be seen that the distribution of the two labels is quite balanced, creating a nice condition for training the model after.

\begin{figure}[H]
  \centering
  \includegraphics[width=0.4\textwidth]{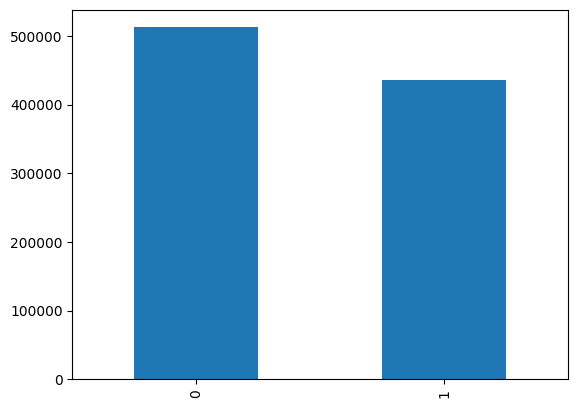}
  \caption{The distribution of labels in training set}
\end{figure}


The node.tsv file contains 837,834 nodes with their Id and text content. However, most of the nodes contain special characters like '{', '|'. '-',... With the method we proposed, the appearance of these characters will negatively affect the accuracy of the POS tagging algorithm. That's why we preprocess the text to replace these characters with spaces.

\begin{figure}[H]
  \centering
  \includegraphics[width=0.4\textwidth]{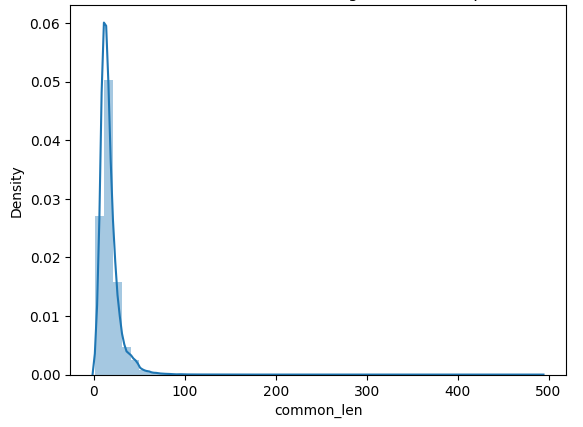}
  \caption{Distribution of common words' length between a pair of nodes}
\end{figure}

Figure 2 shows the distribution of the number of common words in pairs of nodes with a label of 1 in the trainset.


After a general analysis, we do POS tagging for the entire nodeset in the text column with NLTK library. There are total of 37 types of tags that appear in the entire nodeset. Nodeset after being tagged will continue to be analyzed more closely to give meaningful insights.

\begin{figure}[H]
  \centering
  \includegraphics[width=0.5\textwidth]{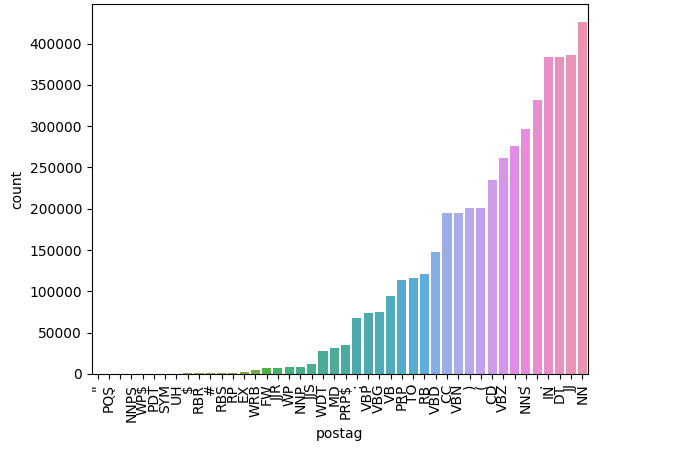}
  \caption{Number of common postag's appearance (not weighted)}
\end{figure}

For each pair of nodes, we get the common tags, and after that the number of appearance of each tag in the entire trainset, in order to extract important insights. Figure 3 shows the number of common tags that appear in the entire train set.

\begin{figure}[H]
  \centering
  \includegraphics[width=0.5\textwidth]{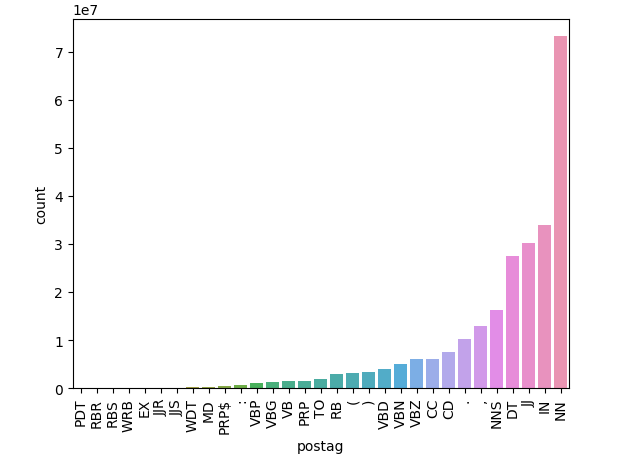}
  \caption{Number of common postag's appearance (weighted)}
\end{figure}

Similar to figure 3, but instead of just getting the common tag that appears in a pair, we get the number of appearances of it. Finally, we sum up the number of times that tag appears in the whole train set and call it weighted. Figure 4 shows the number of common tags that appear in the entire train set.

Besides, based on the number of occurences of the POS tags on the training set, we divided them into 2 group: group 1 includes nodes that have relationship, and group 2 includes nodes with no relationships. We conduct the t-test based on the appearance of each POS tag on the two group to find which set of POS tags have the most impact to the difference between two groups. From our analysis, we found that there is a set of 18 POS tags having the p-value equal to 0.0468. It is clear that with these 18 POS tags, group 1 and group 2 are significantly different. This means these POS tags are potential in detecting the links between nodes. They are \textit{CC, CD, DT, IN, JJ, JJR, JJS, MD, NN, NNS, PRP, PRP\$, RB, TO, VB, VBG, VBN, VBD}. 

\subsection{Link prediction model}
We used various Machine Learning models on the embedded dataset to predict the result. We split the training dataset provided by the competition organizers into the training set and the validation set. We trained the models on the training set and evaluated the results on the validation set. First, we used simple Machine Learning classification algorithms like KNN, Logistic Regression and Decision tree. Through experiments, we found out that the Decision tree has the best prediction result on the validation set. This is reasonable since the tree-based classification algorithms often have competitive results on the classification challenges in the competitions on the websites like Kaggle. Therefore, we began trying out Random Forest, which is an ensemble version of Decision Tree, and this was also the model that produced the best result. We also tried using other tree-based algorithms like Extra trees, XGBoost, LightBGM and Artificial Neural Network (ANN). These models all produced good results but couldn't surpass the Random Forest. We also tried applying Hyperparamter Tuning for the Random Forest classifcation model but the results didn't seem to improve at all. The hyperparameters we used to train the models are listed in Table \ref{hyper} below:
\begin{table}[h]
\begin{center}
\caption{Hyperparameters of Machine Learning models.}
\label{hyper}%
\begin{tabular}{lp{5cm}}
\textbf{Models} & \textbf{Hyperparameters}\\
\hline
SVM & kernel: linear \\
Logistic Regression & solver: lbfgs, multi\_class: multinomial \\
KNN & n\_neighbors: 3 \\
Decision Tree & default\\
Random Forest & n\_estimators: 150 \\
Extra trees & n\_estimators: 150 \\
XGBoost & eta: 0.5, max\_depth: 40, n\_estimators: 250 \\
LightBGM & default\\
ANN & n\_layers: 3, units\_1: 128, act\_1: relu, units\_2: 64, act\_2: relu, units\_3: 32, act\_3: relu, act\_4: sigmoid, optimizer: Adam, learning\_rate: 0.001, loss: binary\_crossentropy\_loss \\
\hline
\end{tabular}
\end{center}
\end{table}
\section{Results}
\subsection{Evaluation of the Node Embedding method}

\begin{table}[ht!]
\centering
\caption{Evaluation of our best model on Public and Private test}
\label{kaggle_result}      
\begin{tabular}{ccc}
\hline
\textbf{Embedding size} & \textbf{Public test} & \textbf{Private test} \\
\noalign{\smallskip}\hline\noalign{\smallskip}
7 tags & 0.99984 & 0.99984 \\
18 tags & 0.99999 & 0.99997 \\
37 tags (full) & 0.99999 & 0.99999 \\
\noalign{\smallskip}\hline\noalign{\smallskip}
\end{tabular}
\end{table}


The table \ref{kaggle_result} compares the results evaluated by the F1-score metric on the public and private test sets for different embedding sizes. We chose these three different embedding sizes for the following reasons: size 7 was visually selected as the first 7 tags in Figure 4 with the highest weights; size 18 was chosen due to t-test analysis in the Data Exploration section; and size 37 was selected as the maximum number of tags appearing in the entire nodeset.


At size 7 and 37, the results measured in the 2 sets are equally good. But at size 18, the result of the private set is slightly reduced. This proves that there is a difference in data distribution between public and private tests. Also, the performance loss of embedding with size 18 could be due to the weakness of traditional machine learning models, leading to overfitted susceptibility. But in general, the machine learning models we used performed very effectively on this dataset. This is partly due to the well-balanced dataset and the good results achieved by feature extraction.


We intend to apply better techniques to solve this problem such as applying deep learning to handle dynamic link prediction,... and conduct comparative testing with other POS Taggers besides NLTK.

\subsection{Evaluation of the Machine Learning models} 
Before submitting the results, we trained the models on the training set and evaluated the results on the validation set first to find the suitable models for the problem. The models are evaluated by using the accuracy and F1-score metrics and the results are shown in table \ref{val_result} below.

\begin{table}[ht!]
\centering
\caption{Evaluation of the Machine Learning models on the validation set.}
\label{val_result}      
\begin{tabular}{ccc}
\hline\noalign{\smallskip}
\textbf{Model} & \textbf{Accuracy} & \textbf{F1-score}  \\
\noalign{\smallskip}\hline\noalign{\smallskip}
Logistic Regression & 0.63 & 0.61 \\
KNN & 0.92 & 0.92 \\
Decision Tree & 0.99988 & 0.99988\\
\textbf{Random Forest} & \textbf{0.99998945} & \textbf{0.99998854} \\
Extra trees & 0.9999894 & 0.99998854 \\
XGBoost & 0.99994727 & 0.99994269 \\
LightBGM & 0.99857631 & 0.99845275 \\
ANN & 0.99828103 & 0.99826962 \\
\noalign{\smallskip}\hline\noalign{\smallskip}
\end{tabular}
\end{table}

As mentioned above, except for the tree-based algorithms and Artificial Neural Network, other Machine Learning models do not have good results for this task. Therefore, we chose the Random Forest, XGboost and Artificial Neural Network models as the final candidates to predict the test results. We didn't select the Extra Trees model since this model has similar result compared to the Random Forest Classifier model, and they are both ensemble version of the Decision Tree classifier. We trained the models on the training set using the nltk full embedding method and submit the predicted test results to the competition. Our submissions are evaluated on the public and private test set using F1-score metric, and the results are shown in table \ref{test_result} below:

\begin{table}[ht!]
\centering
\caption{Evaluation on the public and private test set.}
\label{test_result}      
\begin{tabular}{ccc}
\hline\noalign{\smallskip}
\textbf{Model} & \textbf{Public} & \textbf{Private}  \\
\noalign{\smallskip}\hline\noalign{\smallskip}
\textbf{Random Forest} & \textbf{0.99999} & \textbf{0.99999} \\
XGBoost & 0.99591 & 0.99533 \\
ANN & 0.99994 & 0.99992 \\
\noalign{\smallskip}\hline\noalign{\smallskip}
\end{tabular}
\end{table}

Our proposed methods all achieved excellent results on both the public and private test set. Also, the models' performances on the public and private test set are very consistent, which proves that our models do not overfit the dataset.
\section{Conclusion}
In this paper, we have presented our approach for the link prediction task on Wikipedia articles by using the extracted textual features from the articles through POS Tagging. We explored the dataset by plotting the distribution of labels, word's length and the number of common POS tags' appearances. Through this process, we know that the dataset is quite balanced, which explains for the high prediction results we achieved. The Random Forest got the highest result at 0.99999 evaluated by the F1-score metric on both the public and private test set using the full tag list of 37 tags. The results have proven the efficiency and effectiveness of our solution. Unlike other approaches that only use the id of the node as input, we have utilized the textual features extracted from the articles, so our solution can generalize well and we don't need to train the model again when there is a change in the node id.

\section*{Acknowledgments}
This research was supported by The VNUHCM-University of Information Technology's Scientific Research Support Fund"

\bibliographystyle{IEEEtran}
\bibliography{references}

\end{document}